\documentclass[final]{cvpr}

\usepackage{times}
\usepackage{epsfig}
\usepackage{graphicx}
\usepackage{amsmath}
\usepackage{amssymb}
\usepackage{caption}
\usepackage{subcaption}
\usepackage{booktabs}
\usepackage{comment}
\usepackage{multirow}
\usepackage{placeins}
\usepackage{dblfloatfix}    
\usepackage{float}

\usepackage{makecell}
\setcellgapes{0pt}   

\usepackage[pagebackref=true,breaklinks=true,colorlinks,bookmarks=false]{hyperref}

\def\cvprPaperID{5569} 
\def\confYear{CVPR 2021}

\begin{document}

\title{Robustness Out of the Box: Compositional Representations Naturally Defend Against Black-Box Patch Attacks}

\author{Christian Cosgrove\qquad Adam Kortylewski\qquad Chenglin Yang \qquad Alan Yuille\\
Johns Hopkins University\\
}

\maketitle



\begin{abstract}

Patch-based adversarial attacks introduce a perceptible but localized change to the input that induces misclassification. 
While progress has been made in defending against imperceptible attacks, it remains unclear how patch-based attacks can be resisted.
In this work, we study two different approaches for defending against black-box patch attacks. 
First, we show that adversarial training, which is successful against imperceptible attacks, has limited effectiveness against state-of-the-art location-optimized patch attacks.
Second, we find that compositional deep networks, which have part-based representations that lead to innate robustness to natural occlusion, are robust to patch attacks on PASCAL3D+ and the German Traffic Sign Recognition Benchmark, without adversarial training. Moreover, the robustness of compositional models outperforms that of adversarially trained standard models by a large margin.
However, on GTSRB, we observe that they have problems discriminating between similar traffic signs with fine-grained differences.
We overcome this limitation by introducing part-based finetuning, which improves fine-grained recognition.
By leveraging compositional representations, this is the first work that defends against black-box patch attacks without expensive adversarial training. This defense is more robust than adversarial training and more interpretable because it can locate and ignore adversarial patches.

\end{abstract}

\section{Introduction}
\begin{figure}
    \centering
    \begin{subfigure}{\linewidth}
        \centering
        \includegraphics[width=.9\linewidth]{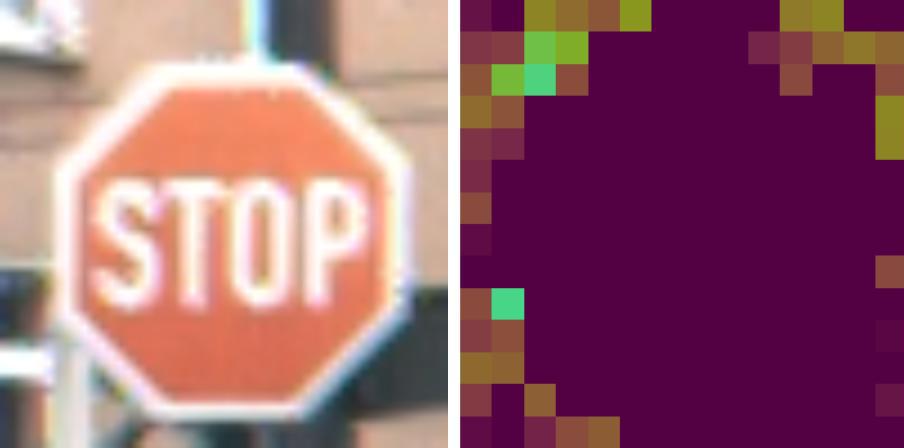}
    \end{subfigure}\\
    \begin{subfigure}{\linewidth}
        \centering
        \includegraphics[width=.9\linewidth]{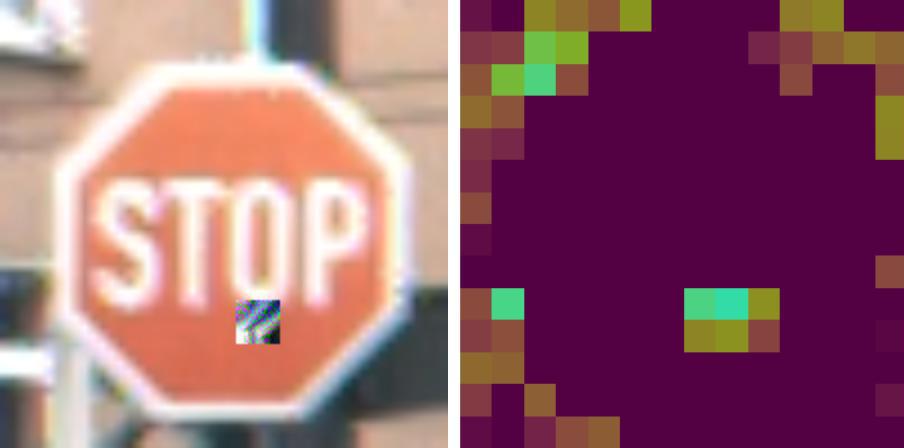}
    \end{subfigure}
    \caption{An attempted Texture Patch Attack~\cite{yang2020patchattack} that fails to induce misclassification. The CompNet has detected the adversarial patch and ignored it. In the right figures, brighter color intensity corresponds to higher \emph{occlusion scores}---regions that the CompNet recognizes as occluders or background.}
    \label{fig:stopsign}
\end{figure}

Patch-based adversarial examples are a powerful class of adversarial attacks, introduced in \cite{brown2017adversarial}. In contrast with gradient-based adversarial examples, these attacks modify the input in a perceptible way but only in a localized region. They are general, and they can be performed in a black-box manner~\cite{yang2020patchattack, croce2020sparsers}, \ie, they do not require access to the parameters of the model.
Patch-based adversarial attacks have been shown to fool state-of-the-art defenses, even in the black-box setting~\cite{yang2020patchattack}.
Reliable defenses against black-box patch attacks remain elusive, and still little is known why deep networks fail to resist adversarial patches. In this work, we study two orthogonal perspectives for defending against black-box patch attacks: \emph{adversarial training} and \emph{deep compositional architectures}.

The first perspective---that adversarial training \cite{madry2017towards_resistant} is a potential solution for protecting against patch attacks---is based on the assumption that deep networks can learn to be invariant to adversarial examples, if those examples are included in the training data. Adversarial training has been shown to be highly effective against imperceptible attacks \cite{madry2017towards_resistant,xie2019denoising}. Moreover, a recent study showed promising results for defending against patch attacks using adversarial training~\cite{rao2020adversarial_patch_training}.

Our extensive experiments confirm that adversarial training, as suggested in \cite{rao2020adversarial_patch_training}, improves the robustness of deep networks to state-of-the-art patch attacks, reducing the success rate to about 80\% for PatchAttack \cite{yang2020patchattack} and 75\% for Sparse-RS \cite{croce2020sparsers} on the PASCAL3D+ dataset. However, in general, the attack success rates remain high: deep networks have difficulties in learning to be invariant to patch-based attacks. The challenging combinatorial variability of patch position and texture remains. Similar limitations of deep networks have also been observed in recent work on training with data augmentation to induce robustness to occlusion~\cite{kortylewski2020compnets_ijcv}. 

From a computer vision perspective, patch attacks can be interpreted as a form of partial occlusion. 
In natural images, objects are frequently occluded by other objects, and robustness to partial occlusion is a long-standing problem in computer vision that has received significant attention \cite{wang2017detecting, zhang2018deepvoting, george2017generative}.
The difference between patch attacks and natural occlusion is that the occluder's position and texture is optimized to fool the model; as such, it is a ``maximally difficult" occluder. In this work, we investigate whether models with state-of-the-art robustness to natural occlusion also have enhanced robustness to patch attacks. 

This paper builds on recent work that introduces compositional deep networks (CompNets).
CompNets have been shown to be highly robust to partial  occlusion~\cite{compnet_wacv,korty2020compnets_cvpr,wang2020robust,kortylewski2020compnets_ijcv}, but little has been studied as to whether these models can resist occlusions whose textures and locations are adversarially selected. This is important, since perceptible but localized patches are a primary way that adversarial examples can be manifested in the real world~\cite{brown2017adversarial}.

Our experiments demonstrate that CompNets have a strong natural robustness to patch-based adversarial attacks. As these models are robust to occlusion by design, we find that CompNets \textbf{do not need to be adversarially trained to be robust} to adversarial patches. Moreover, we find that CompNets are \textbf{significantly more robust than comparable CNN models that are trained with adversarial patches}~\cite{rao2020adversarial_patch_training}. We are the first to defend against black-box adversarial patches without adversarial training.

As part of our empirical studies, we observe that CompNets have trouble differentiating classes that are visually similar, \eg, speed limit signs with different numbers. To improve CompNets' ability to differentiate similar classes, we propose a novel finetuning technique, \emph{part-based finetuning}, which makes CompNets' part features more relevant and class-specific. We find that this method, along with other techniques found in the literature~\cite{compnet_wacv}, improves CompNets' classification accuracy on a fine-grained classification dataset (the German Traffic Sign Recognition Benchmark~\cite{Stallkamp2012gtsrb}). This brings CompNets' accuracy to parity with CNNs trained normally and adversarially, while maintaining to superior robustness to patch attacks.

One of the unique properties of CompNets is their interpretability. Because CompNets build a generative model of each class, they can locate and ignore occluders~\cite{compnet_wacv, korty2020compnets_cvpr}. This allows one to visualize a semantic ``occlusion score" corresponding to blocked regions in the image. We find that this interpretability extends to the adversarial case: \textbf{CompNets can detect adversarial patches and ignore them. As such, their robustness is highly interpretable}. We verify these results both quantitatively and qualitatively.

In summary, we make several important contributions in this work. We show:
\begin{enumerate}
    \item Compositional architectures are robust out of the box.
    \item Adversarial training of standard networks does improve robustness, but is not as effective as compositional architectures at defending against patch attacks.
    \item Combining the outputs of standard architectures and compositional architectures leads to the best accuracy-robustness trade-off. These models achieve a high accuracy for non-attacked images and have enhanced robustness to patch attacks compared to adversarially trained models.
    \item Enhancing compositional representations with part-based finetuning leads to even better performance due to improved fine-grained recognition.
\end{enumerate}

\section{Related work}

\textbf{Adversarial patch attacks and defenses.} It is widely known that standard deep network models suffer when presented with occlusion~\cite{zhu2019robustness, compnet_wacv, korty2020compnets_cvpr}. When a fraction of the image is occluded, CNNs perform poorly, even when human subjects were able to handle such occlusion~\cite{zhu2008unsupervised}. 

Deep networks' fragility under occlusion is not limited to ``natural" examples: their accuracy drops to near zero when presented with carefully crafted adversarial patches. The first work on adversarial patches showed that monochromatic patches with randomized locations can fool networks~\cite{fawzi2016measuring}. This attack was further extended to the white-box setting by using model gradients to optimize the patch texture~\cite{brown2017adversarial}.

Newer black-box adversarial patch attacks improve upon these works by adapting the texture of the patch to improve the success rate with less area~\cite{croce2020sparsers, yang2020patchattack}, and refine the location search strategy using reinforcement learning rather than random search~\cite{yang2020patchattack}. These algorithmic improvements reduce the patch area and number of queries needed to induce a misclassification, and they are able to circumvent even state-of-the-art defenses against perturbation-based attacks~\cite{yang2020patchattack}.

Many defenses against perturbation-based adversarial attacks have been proposed~\cite{xie2019denoising, kannan2018adversarial, goodfellow2014explaining, madry2017towards_resistant}; however, defenses against patch-based attacks are less well studied. Two recent works have adapted adversarial training to the patch attack setting. Chiang \etal~\cite{certified_patch_defenses} show that adversarial training can lead to certified robustness against attacks based on small patches (less than 1\% of image area). Moreover, Rao \etal~\cite{rao2020adversarial_patch_training} use adversarial training to defend against black-box patch attacks.

Both of these approaches rely on adversarial training to improve the the robustness of standard CNN architectures to patch-based attacks. In contrast, we are the first to show that black-box patch-based adversarial attacks can be defended against \emph{without adversarial training} by \emph{using an architecture that is innately robust to occlusion}. This is important from a theoretical standpoint, as we approach network architectures that are inherently robust to attacks. This is also important from a practical standpoint, as adversarial training is expensive, especially for sophisticated threat models~\cite{xie2019denoising}.

\textbf{Robustness to partial occlusion.} Occlusion robustness has been widely studied in the computer vision literature. Like adversarial training, some approaches attempt to make models more robust by augmenting the training data with occluded examples~\cite{yun2019cutmix, devries2017improved}. Others have proposed architectural improvements that allow the model to detect and ignore occluders~\cite{xiao2019tdapnet}.

Compositional models~\cite{geman2002,jin2006,fidler2007,zhu2008,wu2010,kortylewski2019greedy} have been studied as one possible architecture that can naturally detect and ignore occlusion~\cite{kortylewski2017model}. Liao \etal~\cite{liao2016learning} integrate compositionality in CNN models by regularizing their features to represent part-like detectors. Zhang \etal~\cite{zhang2018interpretable} also use part detectors with a localized spatial distribution. Compositional Convolutional Neural Networks extend this approach to use a generative model of part activations~\cite{korty2020compnets_cvpr, compnet_wacv}, allowing the model to explain away occlusions.

Kortylewski \etal~\cite{compnet_wacv} propose a method to combine the output of standard deep networks with compositional networks to improve fine-grained recognition.
However, we are the first to consider an explicit approach that improves CompNets' fine-grained recognition accuracy, and to study their robustness to adversarial patch attacks.

\begin{figure*}[t]
     \centering
     \begin{subfigure}{0.49\linewidth}
         \centering
         \begin{subfigure}{0.32\linewidth}
             \centering
             \includegraphics[width=\linewidth]{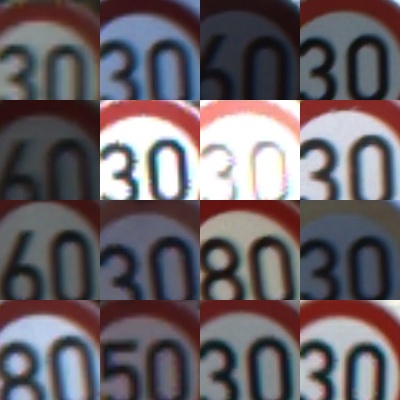}
         \end{subfigure}
         \begin{subfigure}{0.32\linewidth}
             \centering
             \includegraphics[width=\linewidth]{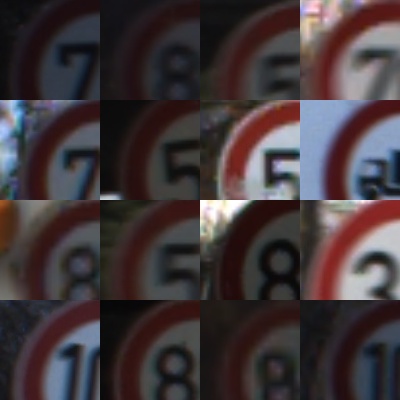}
         \end{subfigure}
         \begin{subfigure}{0.32\linewidth}
             \centering
             \includegraphics[width=\linewidth]{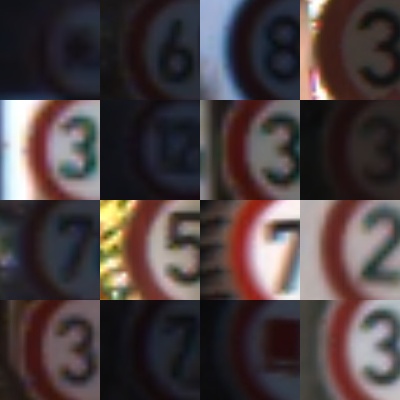}
         \end{subfigure}
         \begin{subfigure}{0.32\linewidth}
             \centering
             \includegraphics[width=\linewidth]{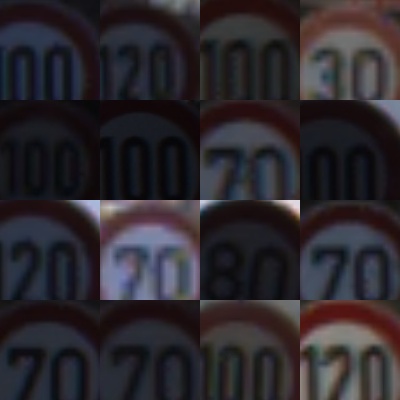}
         \end{subfigure}
         \begin{subfigure}{0.32\linewidth}
             \centering
             \includegraphics[width=\linewidth]{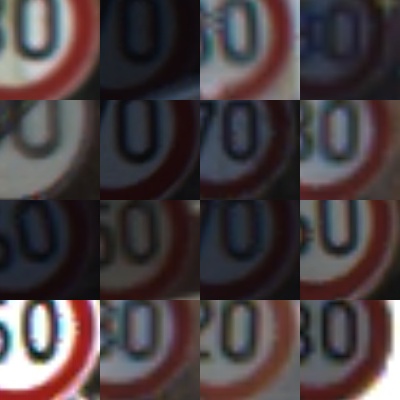}
         \end{subfigure}
         \begin{subfigure}{0.32\linewidth}
             \centering
             \includegraphics[width=\linewidth]{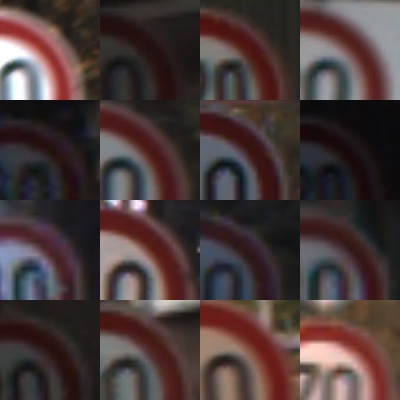}
         \end{subfigure}
         \begin{subfigure}{0.32\linewidth}
             \centering
             \includegraphics[width=\linewidth]{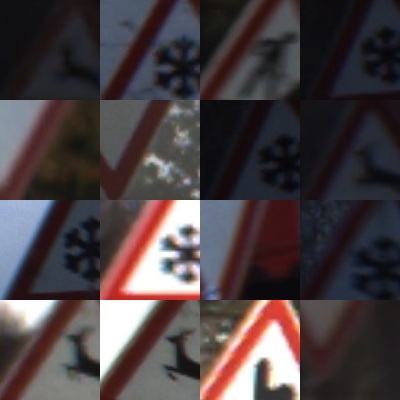}
         \end{subfigure}
         \begin{subfigure}{0.32\linewidth}
             \centering
             \includegraphics[width=\linewidth]{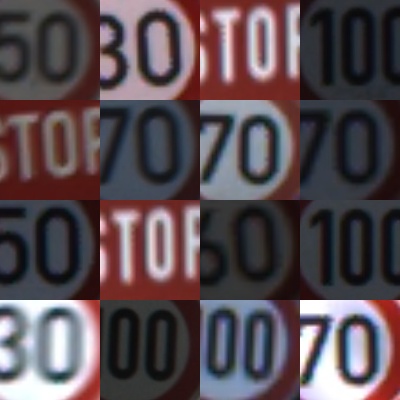}
         \end{subfigure}
         \begin{subfigure}{0.32\linewidth}
             \centering
             \includegraphics[width=\linewidth]{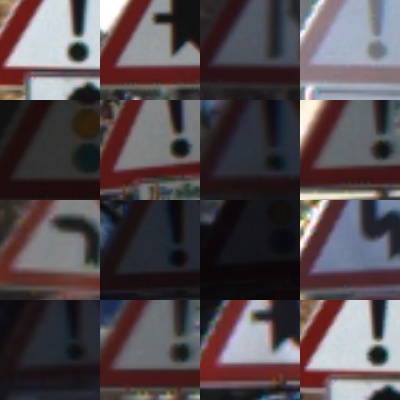}
         \end{subfigure}
        \caption{Without part-based finetuning}
     \end{subfigure}
     \begin{subfigure}{0.49\linewidth}
         \centering
         \begin{subfigure}{0.32\linewidth}
             \centering
             \includegraphics[width=\linewidth]{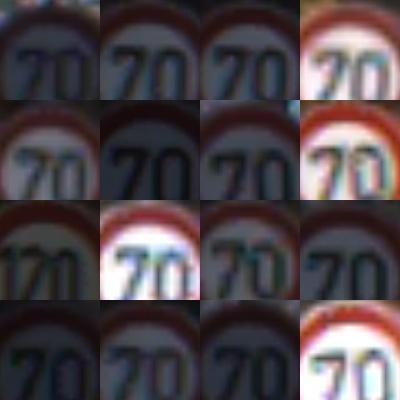}
         \end{subfigure}
         \begin{subfigure}{0.32\linewidth}
             \centering
             \includegraphics[width=\linewidth]{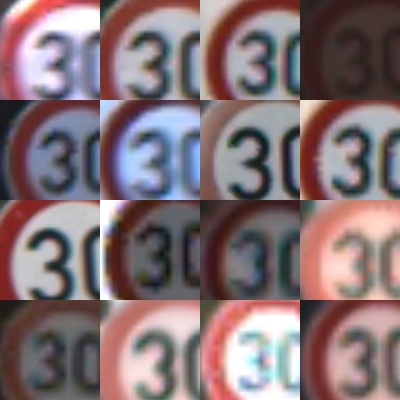}
         \end{subfigure}
         \begin{subfigure}{0.32\linewidth}
             \centering
             \includegraphics[width=\linewidth]{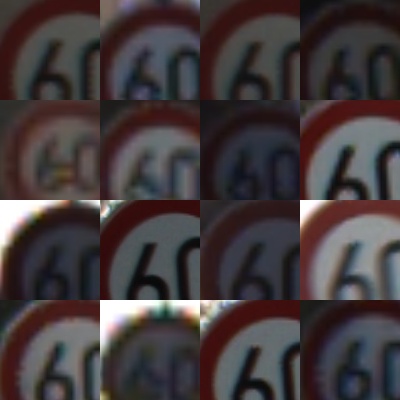}
         \end{subfigure}
         \begin{subfigure}{0.32\linewidth}
             \centering
             \includegraphics[width=\linewidth]{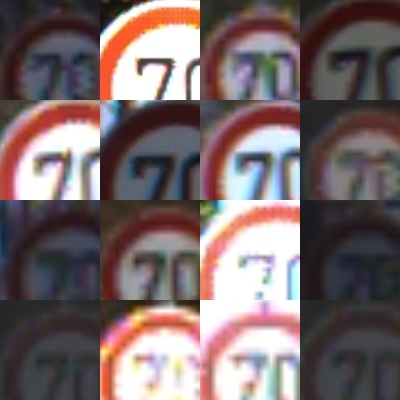}
         \end{subfigure}
         \begin{subfigure}{0.32\linewidth}
             \centering
             \includegraphics[width=\linewidth]{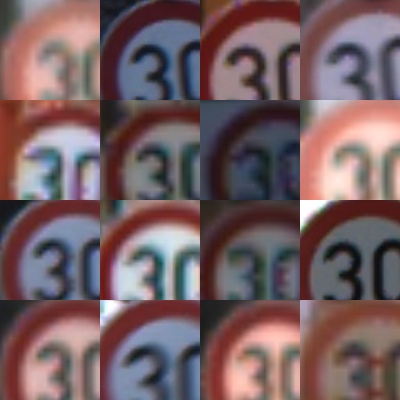}
         \end{subfigure}
         \begin{subfigure}{0.32\linewidth}
             \centering
             \includegraphics[width=\linewidth]{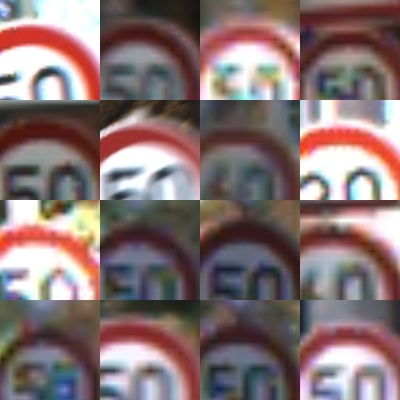}
         \end{subfigure}
         \begin{subfigure}{0.32\linewidth}
             \centering
             \includegraphics[width=\linewidth]{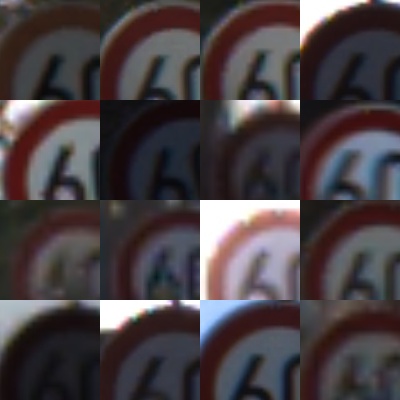}
         \end{subfigure}
         \begin{subfigure}{0.32\linewidth}
             \centering
             \includegraphics[width=\linewidth]{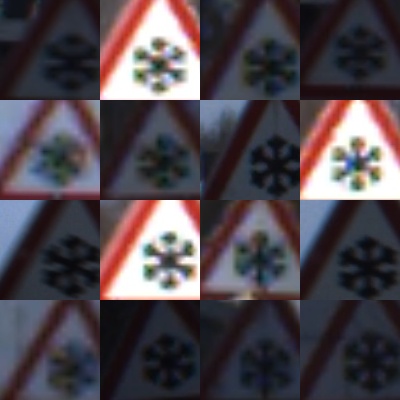}
         \end{subfigure}
         \begin{subfigure}{0.32\linewidth}
             \centering
             \includegraphics[width=\linewidth]{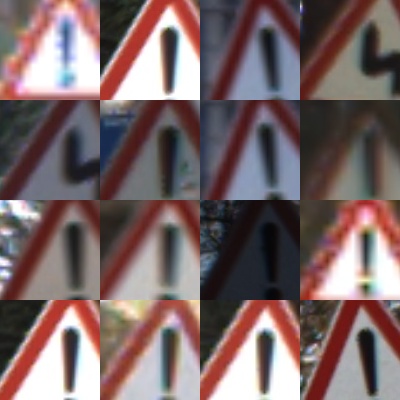}
         \end{subfigure}
        \caption{With part-based finetuning}
     \end{subfigure}
    \caption{Part-based finetuning learns more class-specific and discriminative vMF clusters, increasing CompNets' fine-grained classification performance.}
    \label{fig:part_finetuning}
\end{figure*}

\section{Methods}
\label{sec:methods}

In this section, we explain the prior formulation of CompNets and describe part-based finetuning, our improvement of CompNets for fine-grained classification.
\subsection{Prior work}
\label{sec:prior}
\textbf{Notation.} The output of the layer $l$ in the DCNN is referred to as \textit{feature map} $F^l = \psi(I,\Omega) \in 
\mathbb{R}^{H \times W \times D}$, where $I$ and $\Omega$ are the input image and the parameters of the feature extractor, respectively. 
\textit{Feature vectors} are vectors in the feature map, $f_i^l \in \mathbb{R}^D$ at position $i$, where $i$ is defined on the 2D lattice of $F^l$ with $D$ being the number of channels in the layer $l$ . 
We omit subscript $l$ in the following for clarity since the layer $l$ is fixed \emph{a priori} in the experiments.

\textbf{CompNets.} Compositional Convolutional Neural Networks (CompNets) \cite{korty2020compnets_cvpr} are deep neural network architectures in which the fully connected classification head is replaced with a differentiable compositional model. 
In particular, the classification head defines a probabilistic generative model $p(F|y)$ of the features $F$ for an object class $y$: 
\begin{equation}
\label{eq:vmf1}
    p(F|\Theta_y) = \sum_m \nu_m p(F|\theta^m_y), \hspace{.1cm}\nu_m \in\{0,1\}, \sum_{m=1}^M \nu_m = 1.
\end{equation}
Here, $M$ is the number of mixtures of compositional models per object class and $\nu_m$ is a binary assignment variable that indicates which mixture component is active.
$\Theta_y= \{\theta^m_y = \{\mathcal{A}^m_y,\Lambda\}|m=1,\dots,M\}$ are the overall compositional model parameters for the class $y$. The individual mixture components are defined as:
\begin{equation}
\label{eq:vmf2}
    p(F|\theta^m_y) = \prod_{i} p(f_i|\mathcal{A}_{i,y}^m,\Lambda).
\end{equation}
Note how the distribution \textit{decomposes} the feature map $F$ into a set of individual feature vectors $f_i$.
$\mathcal{A}^m_y=\{\mathcal{A}^m_{i,y}|i \in [H, W] \}$ are the parameters of the mixture components at every position $i$ on the 2D lattice of the feature map $F$. 
In particular, $\mathcal{A}^m_{i,y} = \{\alpha^m_{i,k,y}|k=1,\dots,K\}$ are mixture coefficients and $\Lambda = \{\lambda_k = \{\sigma_k,\mu_k \} | k=1,\dots,K \}$ are the parameters of von-Mises-Fisher (vMF) distributions:
\begin{equation}
\label{eq:vmf3}
    p(f_i|\mathcal{A}_{i,y}^m,\Lambda) = \sum_k \alpha_{i,k,y}^m p(f_i|\lambda_k),
\end{equation}
\begin{equation}
\label{eq:vmfprob}
    p(f_i|\lambda_k) = \frac{e^{\sigma_k \mu_k^T f_i}}{Z(\sigma_k)}, ||f_i|| = 1, ||\mu_k|| = 1.
\end{equation}
Note that $K$ is the number of components in the vMF mixture distributions and  $\sum_{k=0}^K \alpha^m_{i,k,y} = 1$. $Z(\sigma_k)$ is the normalization constant. The model parameters $\{\Omega,\{\Theta_y\}\}$ can be trained end-to-end as discussed in \cite{korty2020compnets_cvpr}.

\textbf{Partial occlusion.}
Compositional networks can be augmented with an outlier model to enhance their robustness to partial occlusion and patch attacks. 
The intuition is that at each position $i$ in the image, either the object model $p(f_i|\mathcal{A}^m_{i,y},\Lambda)$ or an outlier model $p(f_i|\beta,\Lambda)$ is active:
\begin{align}
	&p(F|\theta^m_y,\beta)\hspace{-0.075cm} =\hspace{-0.075cm} \prod_{i} p(f_i|\beta,\Lambda)^{1-z^m_i} p(f_i|\mathcal{A}^m_{i,y},\Lambda)^{z^m_i}.\label{eq:occ}
\end{align}
The binary variables $\mathcal{Z}^m=\{z^m_i \in \{0,1\} | i \in \mathcal{P}\}$ indicate if the object is occluded at position $i$ for mixture component $m$. 
The outlier model is defined as: 
\begin{align}
p(f_i|\beta,\Lambda) = \sum_{k} \beta_{n,k} p(f_i|\sigma_k,\mu_k).
\end{align}
Note that the model parameters $\beta$ are independent of the position $i$ in the feature map and thus the model has no spatial structure. The parameters of the occluder models $\beta$ are learned from clustered features of random natural images that do not contain any object of interest \cite{kortylewski2020compnets_ijcv}. 

When the occlusion model assigns a higher probability (or weight) to a particular region than the class-specific mixture model, the CompNet recognizes an occluder and ignores this region when making a classification decision. This allows one to visualize a spatial map of ``occlusion scores," which provides the basis for CompNets' interpretability (see the end of Section~\ref{sec:experiments}). This occlusion model can also fire for the background, as in Figure~\ref{fig:stopsign}.
\subsection{Part-based finetuning}
\label{sec:part_based_finetuning}
When training on German Traffic Sign dataset, CompNets have trouble differentiating speed limit signs, which share the overall shape and color but differ in their digits. 
To deal with this difficulty, we propose a principled way of improving CompNets' fine-grained classification accuracy.

To make feature vectors more class-specific, we finetune them to be \emph{predictive of the image class}. This causes the CompNet to learn to more directly associate specific features with certain classes, improving fine-grained classification performance. In particular, rather than using a fully-connected layer to synthesize features across the image, we predict the class \emph{directly} from each image feature location using a simple linear classifier:
\begin{equation}
    p(y|f_i) = \text{softmax}(W f_i),
\end{equation}
where $i$ indexes over different local features in the image. However, since CNN features are local, not all regions can be predictive of the class. Rather, our backbone should use the \emph{most predictive} features to determine the final class output. This justifies a max-pooling operation over classification scores:
\begin{align}
    p(y|F) &= \max_i p(y|f_i)\\
    &= \max_i \text{softmax}(W f_i).
\end{align}
Thus, if a feature tends to be associated with many different classes, its class probability will be exceeded by a feature that is more predictive of a particular class.

Part-based finetuning leads to class-specific vMF clusters. In Figure~\ref{fig:part_finetuning}, vMF clusters corresponding specific classes arise: \eg, the ``70" speed limit and for the ``30" speed limit. In contrast, without part-based finetuning, we observe generic clusters that are shared between the different speed limits. It is also worth noting that adding part-based finetuning makes vMF clusters less redundant---specificity leads to fewer duplicated clusters. As we show in Section~\ref{sec:results_main}, part-based finetuning is one key improvement needed to achieve parity in classification accuracy on the fine-grained GTSRB dataset.

\subsection{Combining CNNs with CompNets}
\label{sec:combining_cnns_compnets}
To improve fine-grained recognition accuracy further, we use the combination approach proposed in~\cite{compnet_wacv}. In this method, we first classify the image with a CNN, and only predict with a CompNet if the classifier's confidence drops below a certain threshold. For most of our experiments, we use a confidence threshold of 0.95.
\section{Experiments}
\label{sec:experiments}
\textbf{Datasets.} We test our models on two image recognition datasets, PASCAL3D+~\cite{pascal3dplus}, and the German Traffic Sign Recognition Benchmark (GTSRB)~\cite{Stallkamp2012gtsrb}. The first of these datasets is used as a benchmark for object recognition in prior literature on occlusion-robust models and compositional representations~\cite{compnet_wacv, korty2020compnets_cvpr}, and the latter is used as an example of a dataset where patch-based adversarial examples could fool computer vision systems in the real world~\cite{rao2020adversarial_patch_training}. On the GTSRB data, we test on a 1000-example subset of the test data to accelerate our experiments.

\textbf{Baselines.} On these two datasets, we compare against two baselines: an standard CNN and a CNN that was trained on patch-based adversarial examples~\cite{rao2020adversarial_patch_training}. For both cases, we use a VGG16~\cite{simonyan2014very} model pretrained on ImageNet~\cite{deng2009imagenet} (the same as the backbone used for our CompNet), and we fine-tune using the standard cross-entropy loss and early stopping on the training set. For patch-based adversarial training, we use the best-known state-of-the-art code provided with~\cite{rao2020adversarial_patch_training}. This work shows that training on a weaker adversary with randomized patch locations improves robustness---hence, we use the attacks proposed in this work, rather than the stronger adversaries on which we test the robustness of our trained models. Regardless, training with a stronger adversary like PatchAttack~\cite{yang2020patchattack} would be impractical due to computational costs.

\textbf{Attacks.} 
We study black-box attacks because they are architecture-agnostic and more likely to arise in the real world~\cite{yang2020patchattack}. In particular, we compare the robustness of these models on two state-of-the-art methods: Texture PatchAttack~\cite{yang2020patchattack} and the patch attack version of Sparse-RS~\cite{croce2020sparsers}. Both of these attack methods use patches whose locations and textures are optimized in a black-box fashion (where the objective is to fool the model with the smallest number of queries possible). The main differences between these methods are as follows:
\begin{enumerate}
    \item Texture PatchAttack uses a predefined \emph{texture dictionary} of patches. For TPA, the texture dictionary is generated per-class, where each texture is optimized to elicit some specific output from the model~\cite{yang2020patchattack}. In contrast, Sparse-RS 
    optimizes the patch contents pixel-by-pixel, for the specific image in question.
    \item Texture PatchAttack uses reinforcement learning to optimize both the contents and locations of the texture patches. For each image, an LSTM model is trained on-the-fly to select a patch from the patch dictionary, a region inside that patch to cut out, and a location at which to place the patch in the target image~\cite{yang2020patchattack}. The authors claim that using RL limits the number of model queries needed to achieve a successful attack. On the other hand, Sparse-RS uses a random search to assign the patch locations and pixel values~\cite{croce2020sparsers}.
\end{enumerate}
We evaluate with these two attack methods not only because they are the current state of the art for patch-based adversarial attacks, but also because they use different methods for determining the patch locations and generating the patch textures.

\textbf{Hyperparameters.} For attacks, we use a batch size of 64. For TPA, we use 40 iterations of learning, and for Sparse-RS we use 10000 steps. Where not stated, we use a maximum occlusion area of 10\% for the PASCAL3D+ dataset, and 1\% for the GTSRB dataset (smaller occlusion area since this dataset relies more on smaller details). When adjusting the number of patches, we keep the total occlusion area fixed. On the GTSRB dataset, we resize images to 224x224, as we found that the CompNet had a better initialization with a larger input. For TPA, we learn separate adversarial texture dictionaries for each dataset. We use the official code provided with each of these papers. Our CompNets are trained on pool4 VGG16 features, as in~\cite{korty2020compnets_cvpr, compnet_wacv}. For training the CompNets, we use the publicly available code released in~\cite{korty2020compnets_cvpr}.

\textbf{Metrics.} We measure \emph{attack success rate}, the fraction of correctly classified examples that are fooled by the attack. We study both untargeted and targeted attacks; where not stated, we use untargeted attacks.

\subsection{Results}
\label{sec:results_main}

\begin{table}
\centering
\begin{tabular}{@{}llcccc@{}}
\toprule
                            & \multicolumn{5}{c}{Attack success rates: PASCAL3D+}                                                                                                                               \\ \midrule
                            &                                                                 & \multicolumn{1}{c|}{Acc.}        & \begin{tabular}[c]{@{}c@{}}TPA\\ $(n=1)$\end{tabular} & \begin{tabular}[c]{@{}c@{}}TPA\\ $(n=4)$\end{tabular} & \begin{tabular}[c]{@{}c@{}}Sparse-RS\\ $(n=1)$\end{tabular} \\ \midrule
\parbox[t]{1mm}{\multirow{4}{*}{\rotatebox[origin=c]{90}{untargeted}}} & VGG16                                                           & \multicolumn{1}{c|}{\textbf{98.8}} & 91.6         & 95.4          & 99.6          \\
                            & \begin{tabular}[c]{@{}l@{}}VGG16\\\hspace{2mm}\small{+ adv. train~\cite{rao2020adversarial_patch_training}}\end{tabular} & \multicolumn{1}{c|}{96.0}          & 34.2         & 79.5          & 75.4          \\
                            & CompNet                                                         & \multicolumn{1}{c|}{98.2}          & \textbf{7.8} & \textbf{24.9} & \textbf{18.0} \\ \midrule
\parbox[t]{1mm}{\multirow{4}{*}{\rotatebox[origin=c]{90}{targeted}}}   & VGG16                                                           & \multicolumn{1}{c|}{\textbf{98.8}} & 52.6         & 88.0          & 84.7          \\
                            & \begin{tabular}[c]{@{}l@{}}VGG16\\\hspace{2mm}\small{+ adv. train~\cite{rao2020adversarial_patch_training}}\end{tabular} & \multicolumn{1}{c|}{96.0}          & 8.6          & 53.3          & 33.5          \\
                            & CompNet                                                         & \multicolumn{1}{c|}{98.2}          & \textbf{2.4} & \textbf{8.2}  & \textbf{5.8}  \\ \bottomrule
\end{tabular}
\caption{CompNets are significantly more robust than normal and adversarially trained CNNs under targeted and untargeted Texture Patch Attacks~\cite{yang2020patchattack} and Sparse-RS attacks~\cite{croce2020sparsers}.}
\label{table:pascal_all}
\end{table}
\begin{table}
\centering
\begin{tabular*}{\linewidth}{@{}lccc@{}}
\toprule
\multicolumn{4}{c}{Untargeted attack success rates: GTSRB}                                                                                                                                                      \\ \midrule         
                            & \multicolumn{1}{c|}{Acc.}        & \begin{tabular}[c]{@{}c@{}}TPA\\ $(n=1)$\end{tabular} & \begin{tabular}[c]{@{}c@{}}Sparse-RS\\ $(n=1)$\end{tabular} \\ \midrule
VGG16                                                                                                                                   & \multicolumn{1}{c|}{95.2}   & 90.1    & 92.6          \\
\begin{tabular}[c]{@{}l@{}}VGG16\\\hspace{2mm}\small{+ adv. train~\cite{rao2020adversarial_patch_training}}\end{tabular}                                                                         & \multicolumn{1}{c|}{95.5}     & 79.9    & 79.4          \\
CompNet                                                                                                                                 & \multicolumn{1}{c|}{61.7}   & \textbf{23.7}    & \textbf{43.9}          \\
\begin{tabular}[c]{@{}l@{}}CompNet\\\hspace{2mm}\small{+ part-based finetuning}\end{tabular}                                                               & \multicolumn{1}{c|}{75.9}   & 40.6    & 64.2          \\
\begin{tabular}[c]{@{}l@{}}CompNet\\\hspace{2mm}\small{+ part-based finetuning}\\\hspace{2mm}\small{+ two-stage combination~\cite{compnet_wacv}}\\\hspace{2mm}\small{threshold=0.99, temperature 2}\end{tabular} & \multicolumn{1}{c|}{85.1}   & 36.4    & 66.0          \\
\begin{tabular}[c]{@{}l@{}}CompNet\\\hspace{2mm}\small{+ part-based finetuning}\\\hspace{2mm}\small{+ two-stage combination~\cite{compnet_wacv}}\\\hspace{2mm}\small{threshold=0.95, temperature 1}\end{tabular}     & \multicolumn{1}{c|}{\textbf{93.0}}   & 64.4    & 75.8          \\ \bottomrule
\end{tabular*}
\caption{CompNets are significantly more robust than normal and adversarially trained CNNs on the GTSRB dataset. Adding our part-based finetuning (Section~\ref{sec:part_based_finetuning}) and two-stage combination~\cite{compnet_wacv} improves accuracy while maintaining CompNets' robustness advantage.}
\label{table:gtsrb_untargeted}
\end{table}
\begin{table}
\centering
\begin{tabular}{llll}
\toprule
\multicolumn{4}{c}{Untargeted TPA success rate: PASCAL3D+} \\ \midrule
\# patches: & $n=1$    & $n=4$    & $n=8$    \\ \midrule
VGG16               & 91.6 & 95.4 & 94.1 \\
VGG16 (+adv. train) & 34.2 & 79.5 & 95.7 \\
CompNet             & \textbf{7.8} & \textbf{24.9} & \textbf{49.2} \\ \bottomrule
\end{tabular}
\caption{CompNets are more robust than adversarially trained CNNs, even with more patches.}
\label{table:num_patch_ablation}
\end{table}
\begin{table}
\centering
\begin{tabular}{llll}
\toprule
\multicolumn{4}{c}{Untargeted TPA success rate: PASCAL3D+} \\ \midrule
occlusion area: & 1\%    & 10\%   & 50\%    \\ \midrule
VGG16               & 33.6 & 91.6 & 100.0 \\
VGG16 (+adv. train) & 13.1 & 34.2 & 96.1  \\
CompNet             & \textbf{2.7} & \textbf{7.8} & \textbf{71.8} \\ \bottomrule
\end{tabular}
\caption{CompNets are more robust than adversarially trained CNNs, even with larger occlusion area.}
\label{table:area_ablation}
\end{table}

\textbf{CompNets are robust to patch attacks.} Tables~\ref{table:pascal_all} and~\ref{table:gtsrb_untargeted} show that that CompNets are robust to patch attacks. When only one patch is used, CompNets are able to defend against more than 90\% of TPA attacks and 80\% of Sparse-RS attacks on PASCAL3D+, and more than 50\% of attacks on GTSRB. Generally, we find that CompNets are more robust on PASCAL3D+ than on GTSRB, which has a larger number of similar classes. This shows that CompNets are the first architecture that is naturally robust to black-box patch attacks.

\textbf{CompNets are more robust than adversarially trained architectures.} Our results show that CompNets are significantly more robust than normal and adversarially trained CNNs on both PASCAL3D+ (Table~\ref{table:pascal_all}) and GTSRB (Table~\ref{table:gtsrb_untargeted}). For example, CompNets are up to 4x more robust than a comparable adversarially trained CNN on PASCAL3D+, and up to 3x more robust on GTSRB. Remarkably, training the CompNet comes at negligible computational cost compared to adversarial training, and it has superior robustness. This result has never been shown before in prior work.

\textbf{Part-based finetuning (Section~\ref{sec:part_based_finetuning}) improves fine-grained recognition.} On the GTSRB traffic sign dataset, the accuracy of a normal CompNet is unsatisfactory. This is due to failures in fine-grained classification. Adding part-based finetuning improves accuracy by nearly 15 points, and using the ensembling approach presented in~\cite{compnet_wacv} improves the accuracy to 93\%, on par with a standard deep network backbone. The best of these models is still more robust than an adversarially trained standard network under these attacks.

\textbf{Trading accuracy for robustness.} It is worth noting that the improvements in fine-grained recognition come at the expense of slightly worse robustness. Adding part-based finetuning decreases robustness slightly, and combining the output of the model with a standard deep network classifier (as in Section~\ref{sec:combining_cnns_compnets}) harms robustness further. This trade-off between accuracy and robustness has been studied before~\cite{zhang2019theoretically}. Interestingly, the adversarially trained model seems to perform approximately as well as the non-adversarially trained model, suggesting that this trade-off is not at play here. This observation is supported by other works~\cite{xie2020adversarial}.

Fine-grained recognition accuracy with part-based finetuning is still lower than ideal, and combining the output of a CompNet with the output of a standard deep network trades accuracy for robustness. We leave open the question of how to improve CompNets' fine-grained recognition performance to future work.









\begin{figure*}[b]
    \centering
    \begin{subfigure}[b]{0.40\linewidth}
        \begin{subfigure}[b]{\linewidth}
            \centering
            \includegraphics[width=\linewidth]{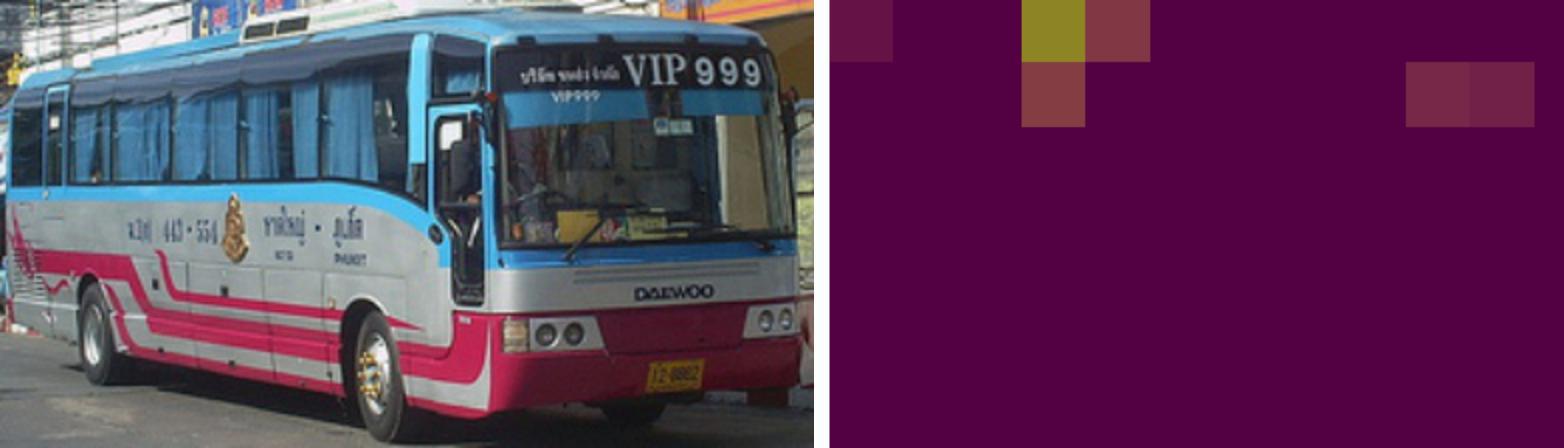}
        \end{subfigure}\\
        \begin{subfigure}[b]{\linewidth}
            \centering
            \includegraphics[width=\linewidth]{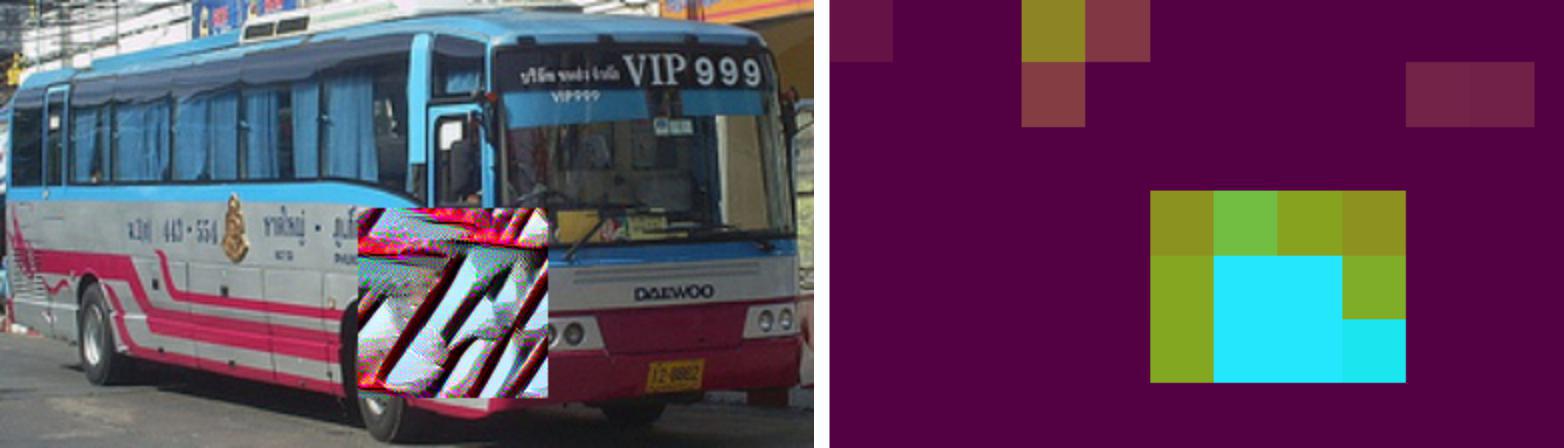}
        \end{subfigure}
        \caption{}
        \label{fig:bus}
     \end{subfigure}
    \begin{subfigure}[b]{0.40\linewidth}
        \begin{subfigure}[b]{\linewidth}
            \centering
            \includegraphics[width=\linewidth]{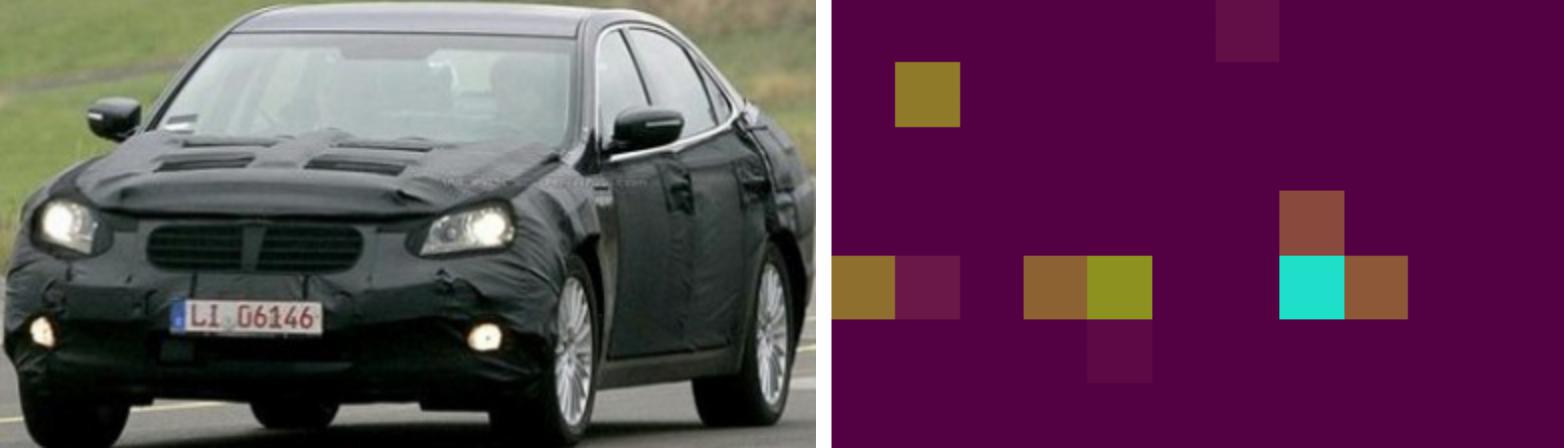}
        \end{subfigure}\\
        \begin{subfigure}[b]{\linewidth}
            \centering
            \includegraphics[width=\linewidth]{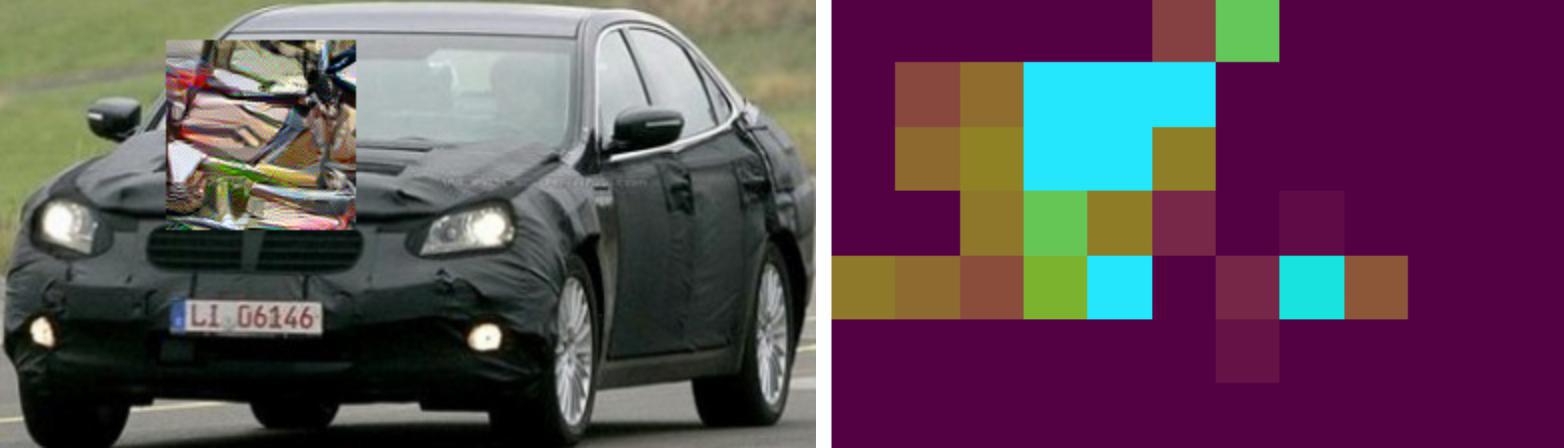}
        \end{subfigure}
        \caption{}
        \label{fig:car}
     \end{subfigure}
    \begin{subfigure}[b]{0.078\linewidth}
        \begin{subfigure}[b]{\linewidth}
            \centering
            \includegraphics[width=\linewidth]{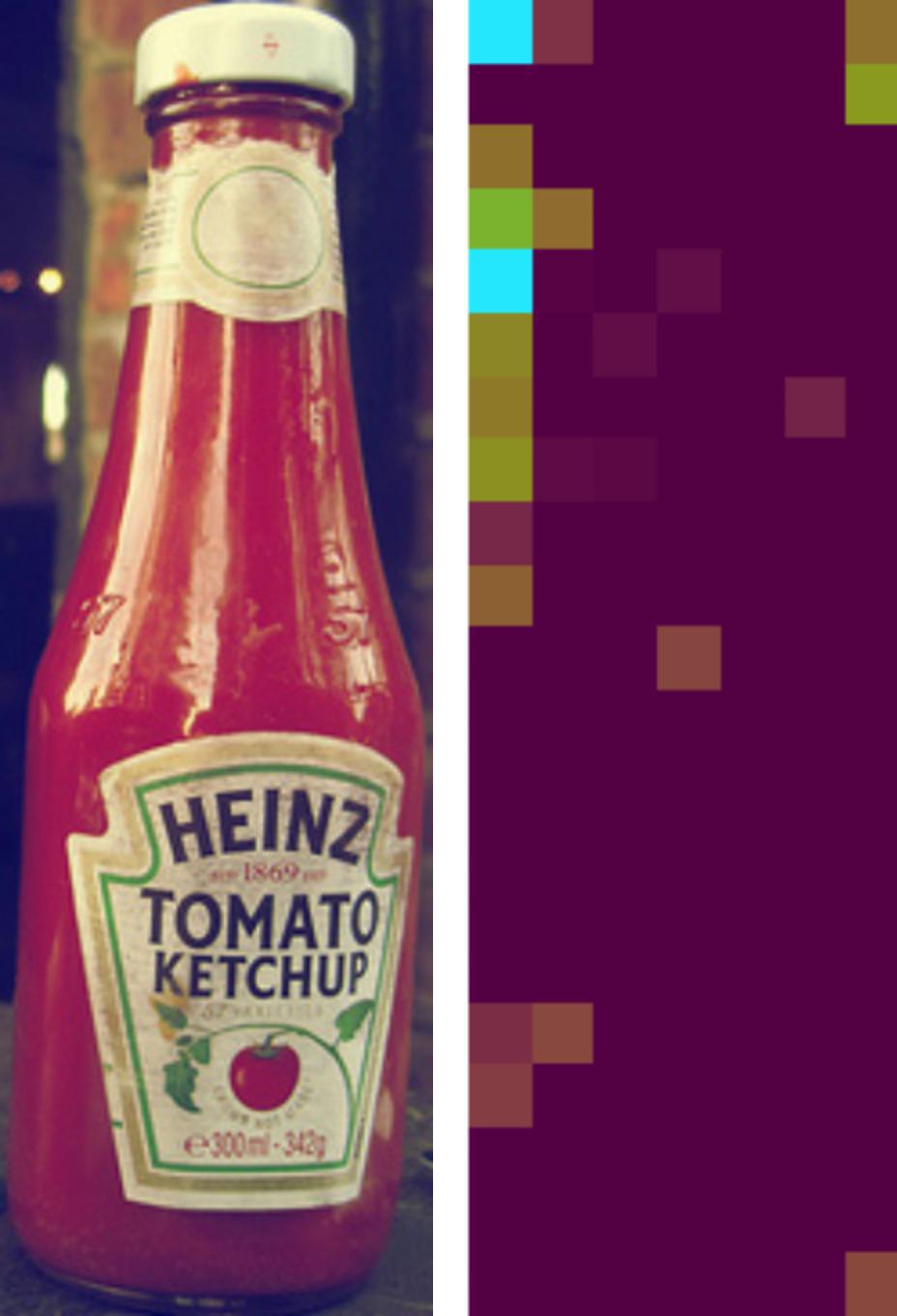}
        \end{subfigure}\\
        \begin{subfigure}[b]{\linewidth}
            \centering
            \includegraphics[width=\linewidth]{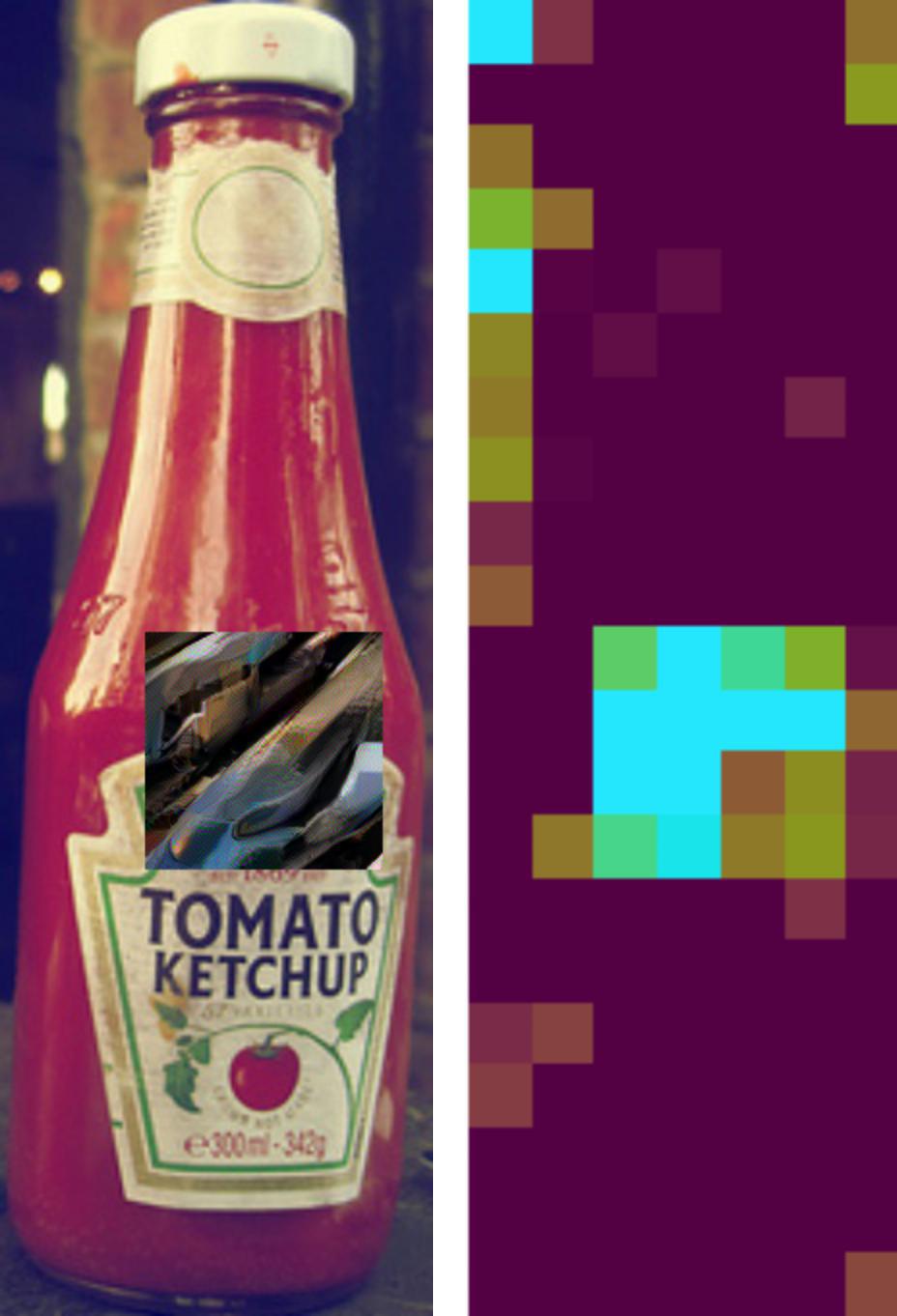}
        \end{subfigure}
        \caption{}
        \label{fig:bottle}
     \end{subfigure}

    \begin{subfigure}[b]{0.46\linewidth}
        \begin{subfigure}[b]{\linewidth}
            \centering
            \includegraphics[width=\linewidth]{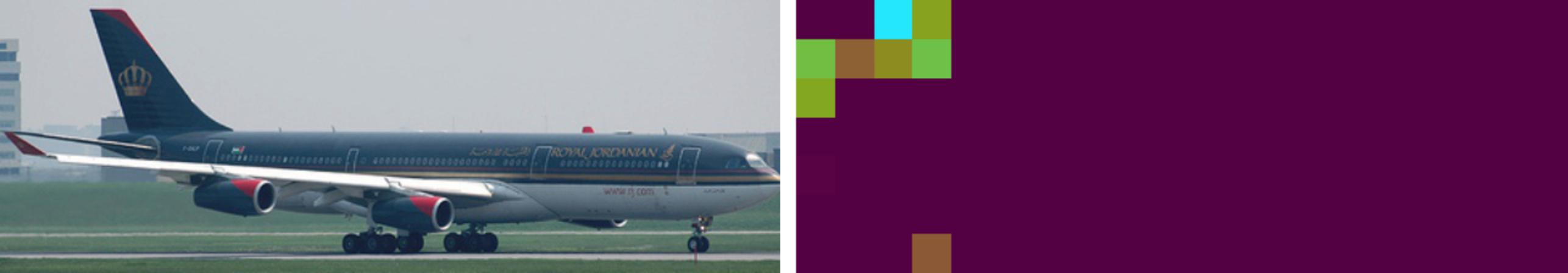}
        \end{subfigure}\\
        \begin{subfigure}[b]{\linewidth}
            \centering
            \includegraphics[width=\linewidth]{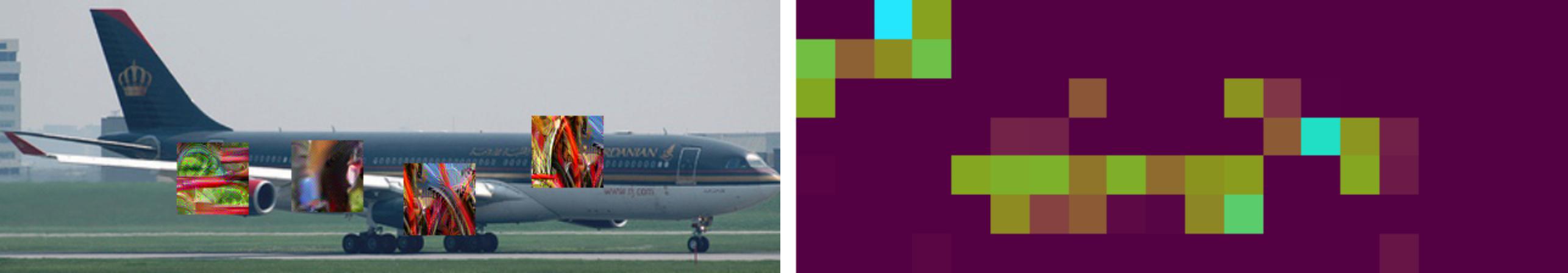}
        \end{subfigure}
        \caption{}
        \label{fig:plane}
     \end{subfigure}
    \begin{subfigure}[b]{0.53\linewidth}
        \begin{subfigure}[b]{\linewidth}
            \centering
            \includegraphics[width=\linewidth]{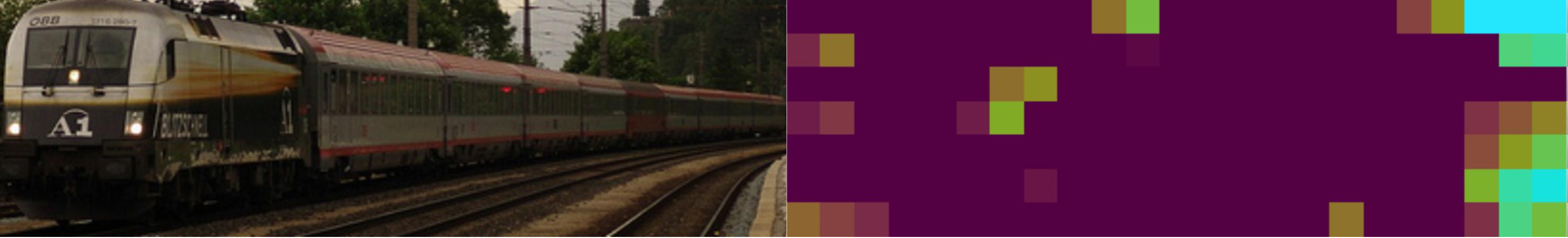}
        \end{subfigure}\\
        \begin{subfigure}[b]{\linewidth}
            \centering
            \includegraphics[width=\linewidth]{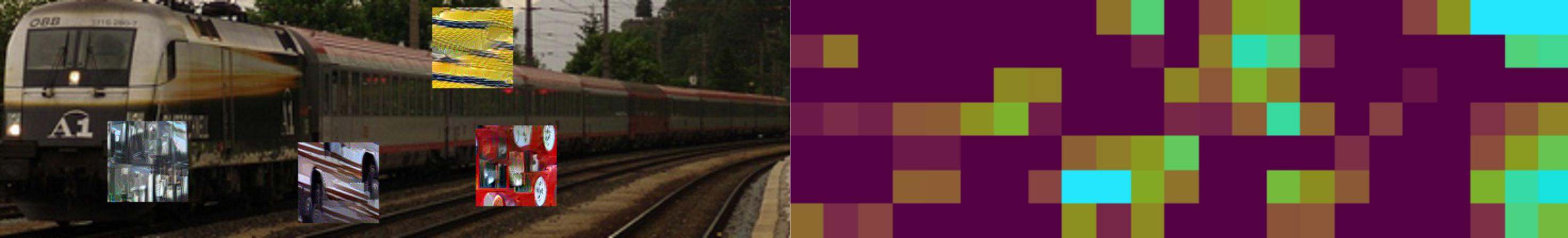}
        \end{subfigure}
        \caption{}
        \label{fig:train}
    \end{subfigure}

    \begin{subfigure}[b]{0.37\linewidth}
        \begin{subfigure}[b]{\linewidth}
            \centering
            \includegraphics[width=\linewidth]{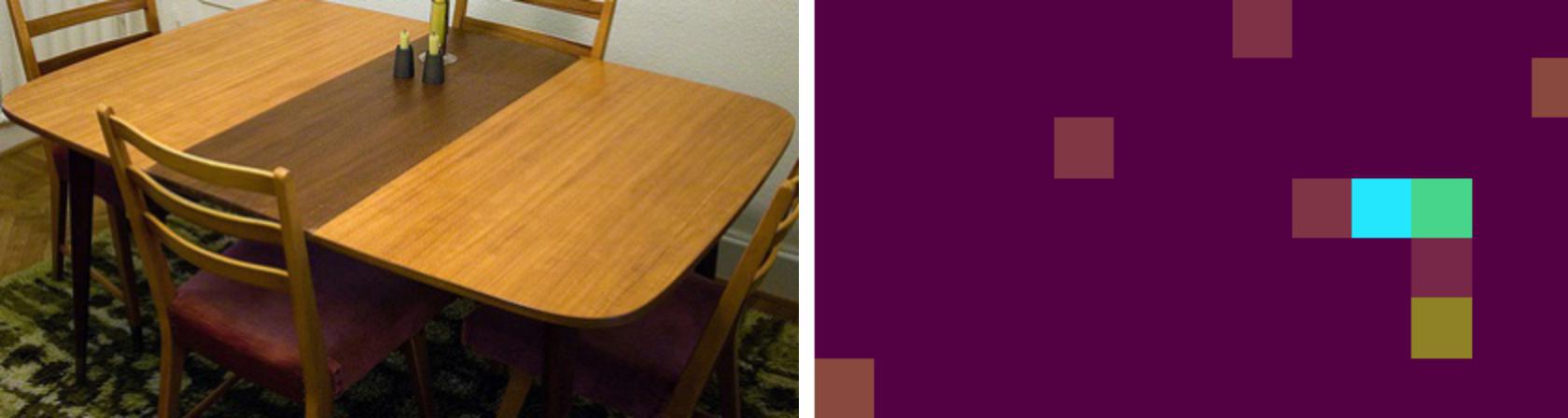}
        \end{subfigure}\\
        \begin{subfigure}[b]{\linewidth}
            \centering
            \includegraphics[width=\linewidth]{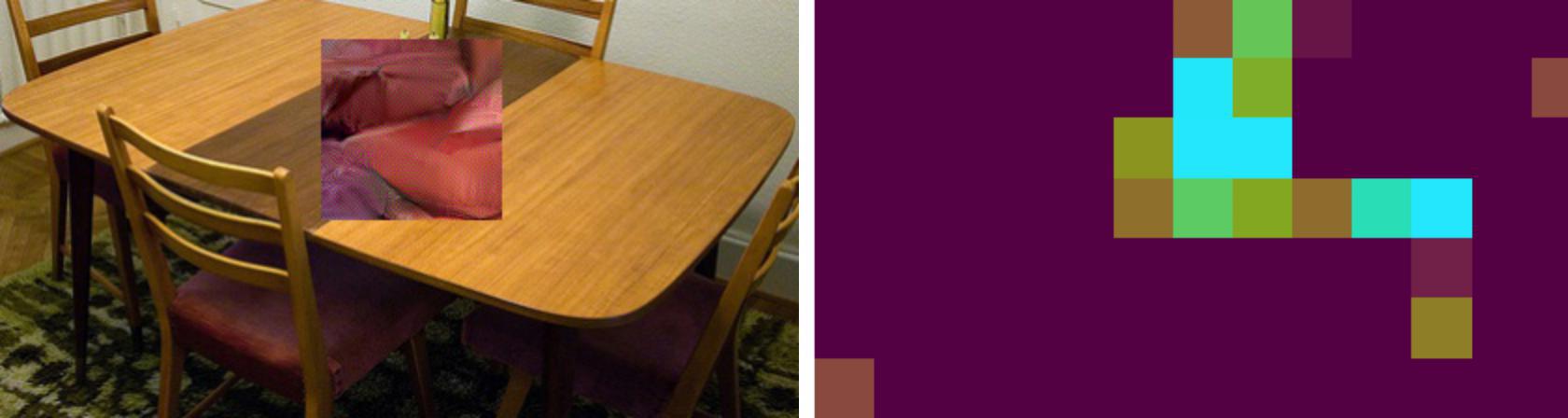}
        \end{subfigure}
        \caption{}
        \label{fig:table}
    \end{subfigure}
    \begin{subfigure}[b]{0.20\linewidth}
        \begin{subfigure}[b]{\linewidth}
            \centering
            \includegraphics[width=\linewidth]{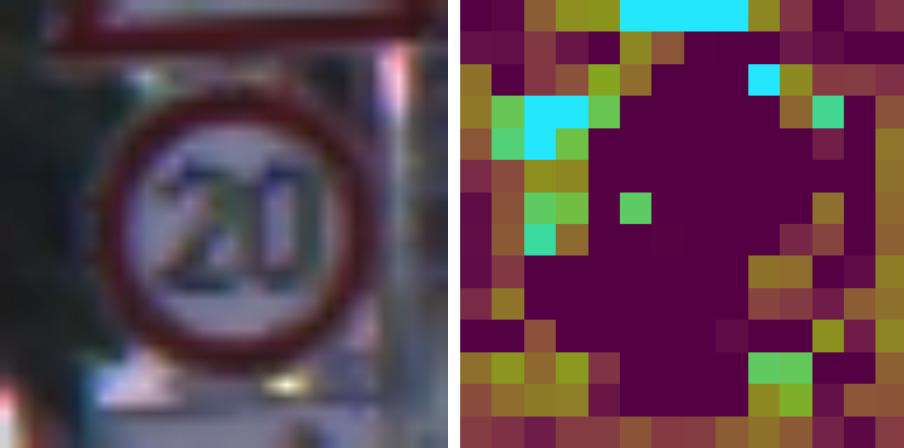}
        \end{subfigure}\\
        \begin{subfigure}[b]{\linewidth}
            \centering
            \includegraphics[width=\linewidth]{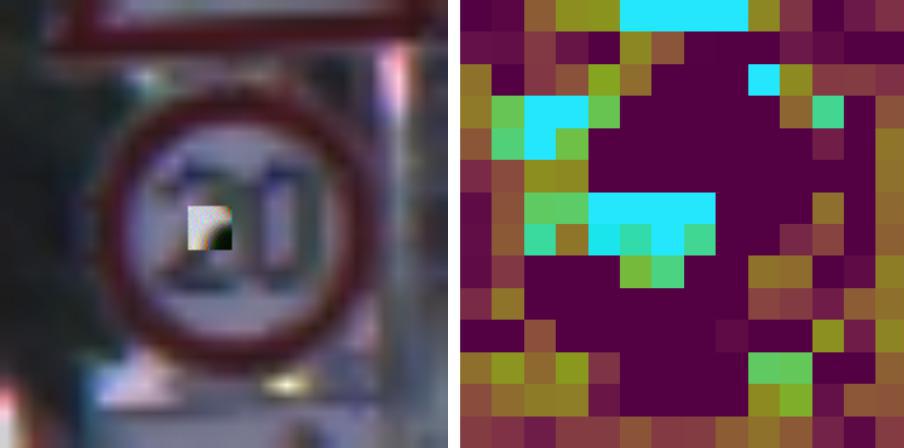}
        \end{subfigure}
        \caption{}
        \label{fig:car}
    \end{subfigure}
     \begin{subfigure}[b]{0.20\linewidth}
        \begin{subfigure}[b]{\linewidth}
            \centering
            \includegraphics[width=\linewidth]{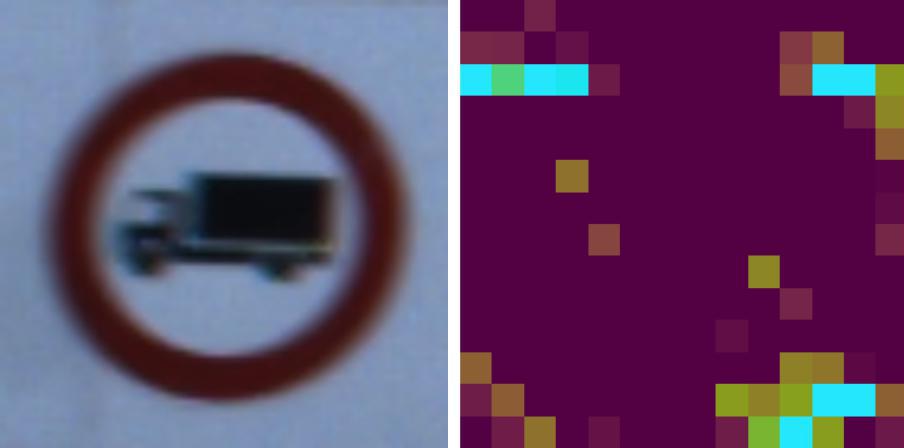}
        \end{subfigure}\\
        \begin{subfigure}[b]{\linewidth}
            \centering
            \includegraphics[width=\linewidth]{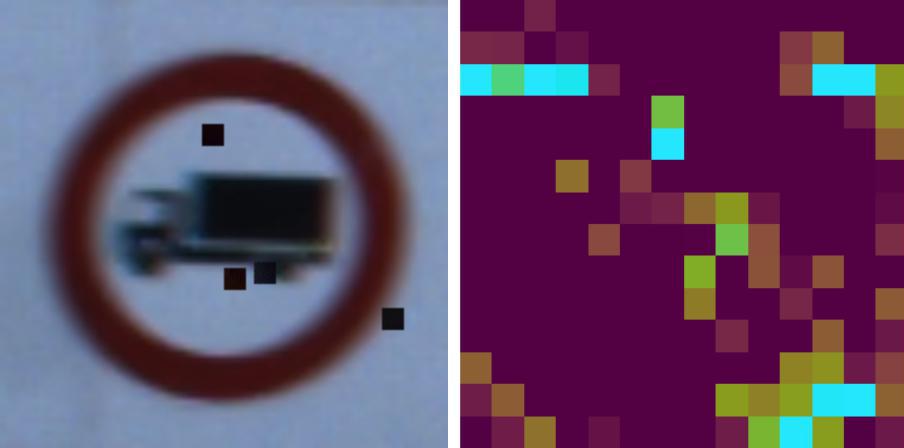}
        \end{subfigure}
        \caption{}
        \label{fig:car}
     \end{subfigure}
     
    \caption{Successful TPA defenses and occlusion maps. Figures~(a-f) are from PASCAL3D+ and~(g,h) are from GTSRB.}
    \label{fig:successful_defenses}
\end{figure*}
\textbf{CompNets are robust under harder attacks.} We conduct two ablation studies: varying the number of patches (Table~\ref{table:num_patch_ablation}) and the occlusion area (Table~\ref{table:area_ablation}). As expected, increasing the number of patches or occlusion area leads to drops in robustness. However, CompNets are able to handle multiple patches more gracefully than the other models: going from one patch to four patches only results in a 17-point increase in attack success rate, whereas the adversarially trained model suffers a 45-point increase in attack success rate. Similar trends can be observed when increasing area: going from 1\% occlusion area to 10\% occlusion area, CompNets' robustness only decreases by 5 points, whereas the adversarially trained CNN takes a 21-point hit to robustness. Overall, CompNets show greater robustness across the board, even against harder attack configurations.


\textbf{CompNets' robustness is interpretable.} Improved robustness to patch attacks is not the only benefit of CompNets to adversarially trained CNNs. CompNets' robustness is highly interpretable: we can visualize which image regions the CompNet recognizes as occluded and explain why the CompNet is able to resist attacks. To locate occluders, we measure the \emph{occlusion scores}: the score (or unnormalized log-probability) that the CompNet assigns to the occlusion model, if the probability of occlusion exceeds some threshold (see Section~\ref{sec:prior}). 

\begin{figure}
    \centering
    \includegraphics[width=\linewidth]{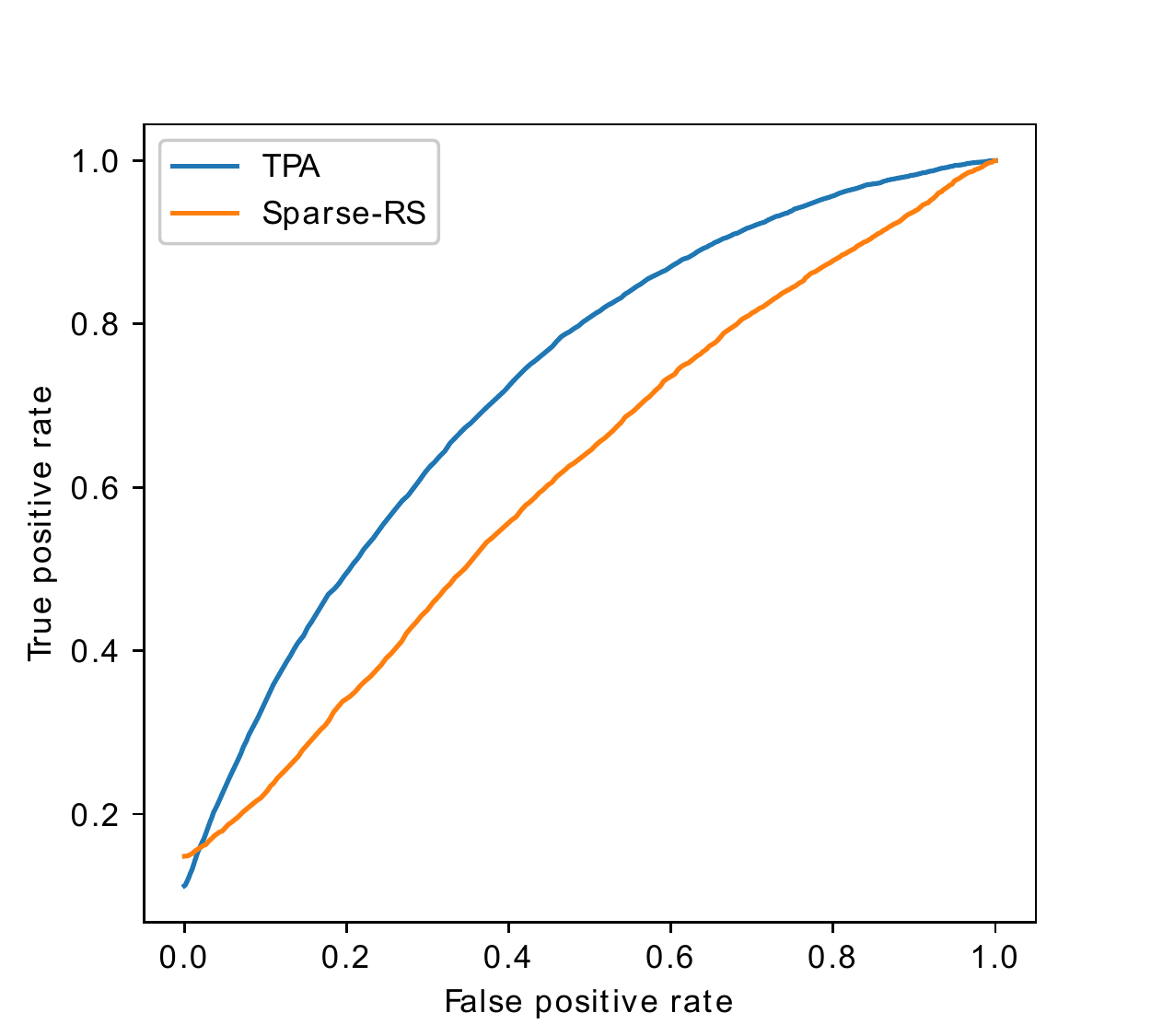}
    \caption{ROC curve for adversarial patch localization on PASCAL3D+.}
    \label{fig:roc}
\end{figure}

\begin{figure*}
    \centering
    \begin{subfigure}[b]{0.352\linewidth}
        \begin{subfigure}[b]{\linewidth}
            \centering
            \includegraphics[width=\linewidth]{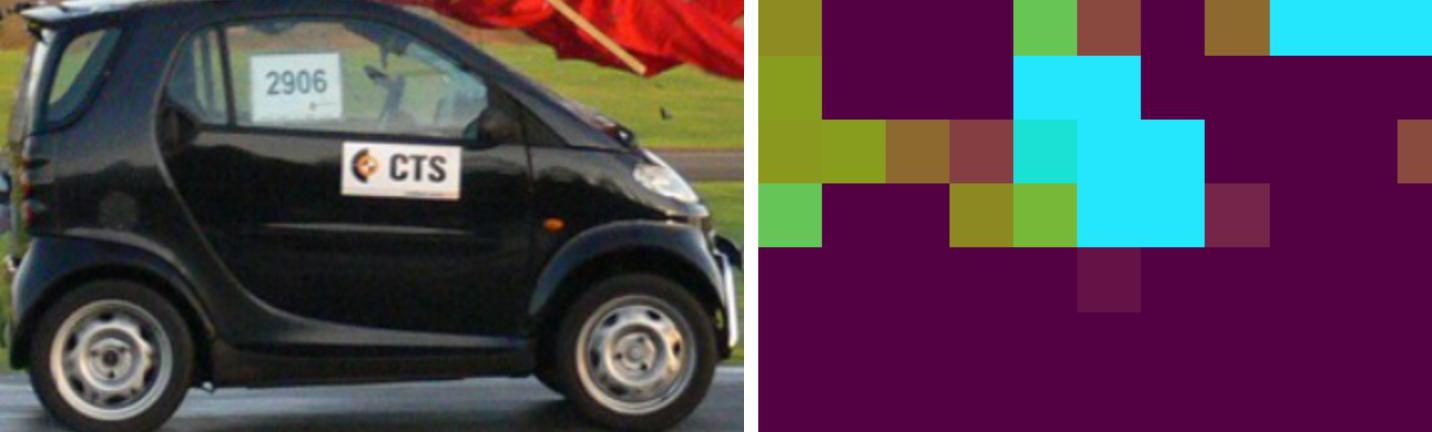}
        \end{subfigure}\\
        \begin{subfigure}[b]{\linewidth}
            \centering
            \includegraphics[width=\linewidth]{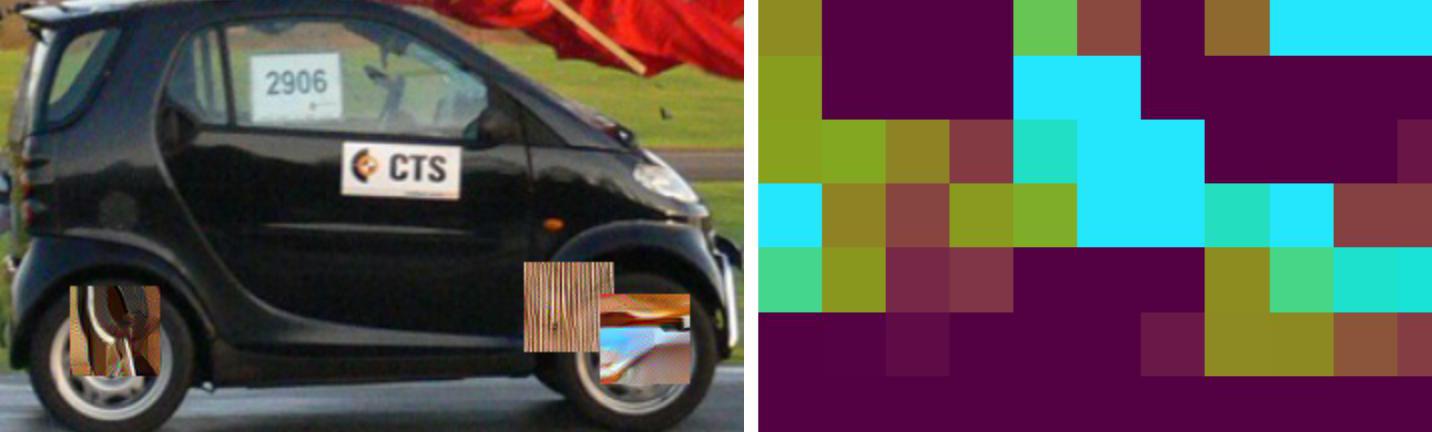}
        \end{subfigure}
        \caption{}
        \label{fig:defense_failure_1}
     \end{subfigure}
    \begin{subfigure}[b]{0.54\linewidth}
        \begin{subfigure}[b]{\linewidth}
            \centering
            \includegraphics[width=\linewidth]{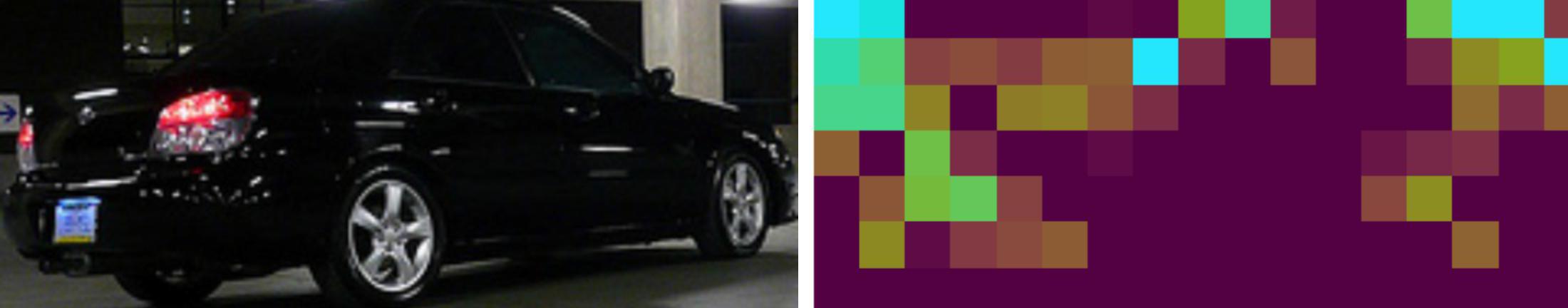}
        \end{subfigure}\\
        \begin{subfigure}[b]{\linewidth}
            \centering
            \includegraphics[width=\linewidth]{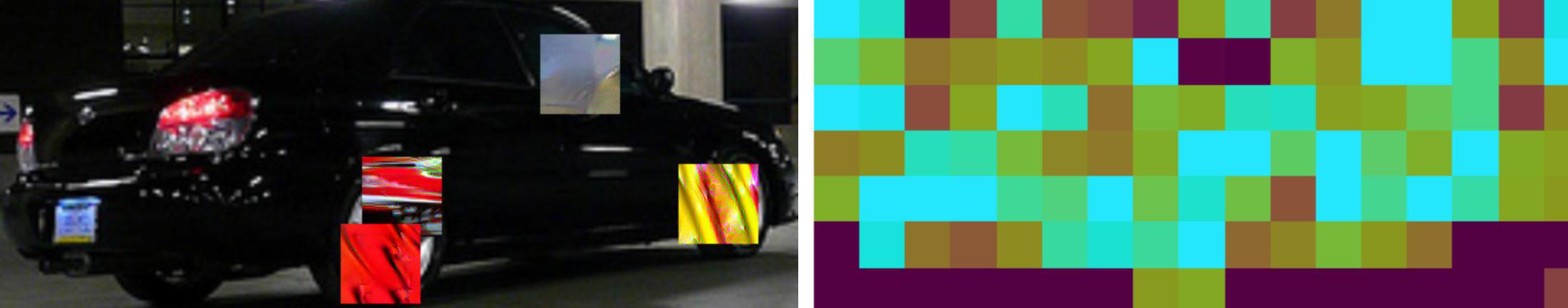}
        \end{subfigure}
        \caption{}
        \label{fig:defense_failure_2}
     \end{subfigure}
     
         \begin{subfigure}[b]{0.22\linewidth}
        \begin{subfigure}[b]{\linewidth}
            \centering
            \includegraphics[width=\linewidth]{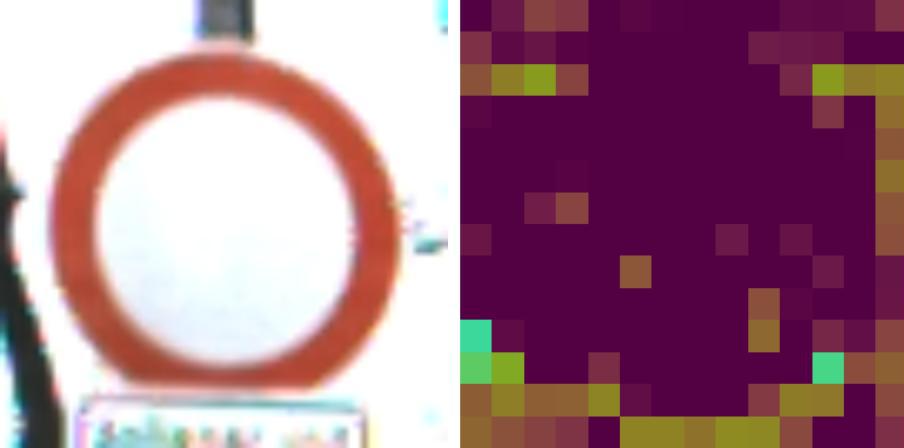}
        \end{subfigure}\\
        \begin{subfigure}[b]{\linewidth}
            \centering
            \includegraphics[width=\linewidth]{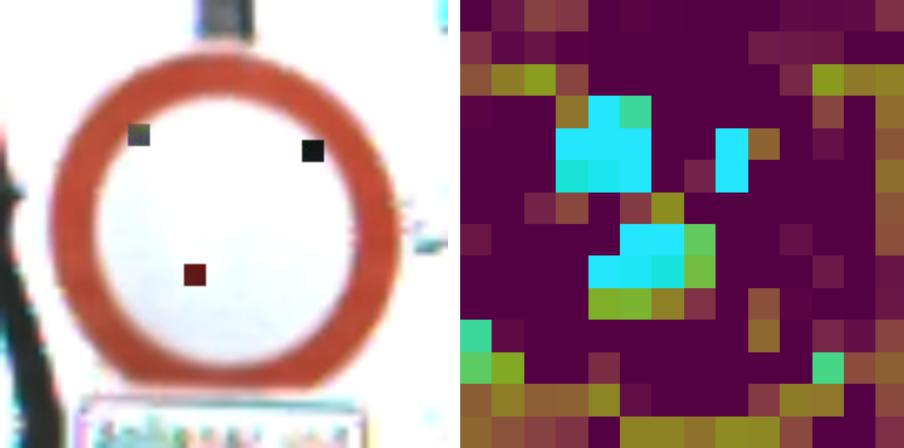}
        \end{subfigure}
        \caption{}
        \label{fig:defense_failure_3}
     \end{subfigure}
    \begin{subfigure}[b]{0.22\linewidth}
        \begin{subfigure}[b]{\linewidth}
            \centering
            \includegraphics[width=\linewidth]{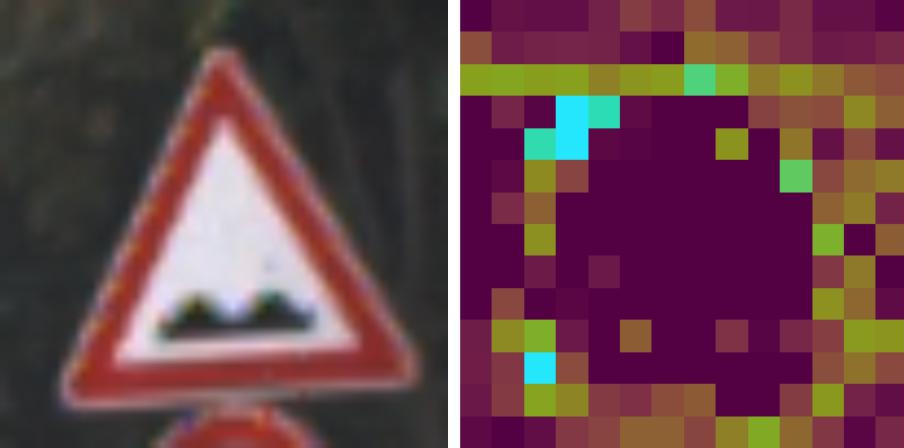}
        \end{subfigure}\\
        \begin{subfigure}[b]{\linewidth}
            \centering
            \includegraphics[width=\linewidth]{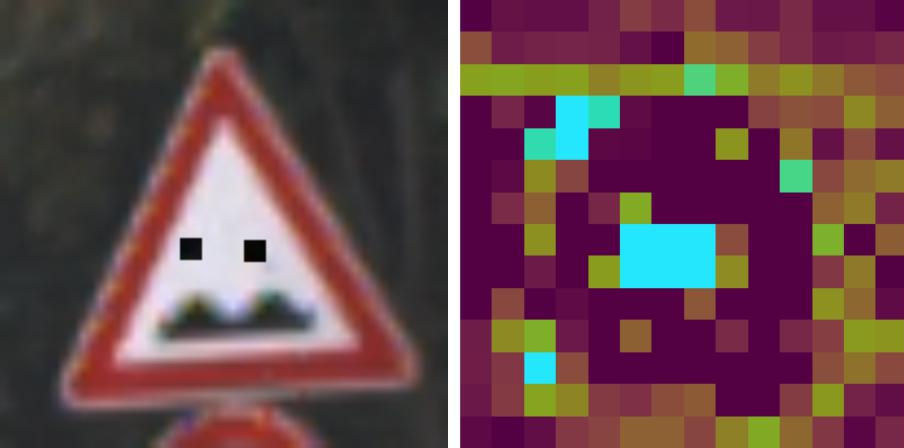}
        \end{subfigure}
        \caption{}
        \label{fig:defense_failure_4}
     \end{subfigure}
    \begin{subfigure}[b]{0.22\linewidth}
        \begin{subfigure}[b]{\linewidth}
            \centering
            \includegraphics[width=\linewidth]{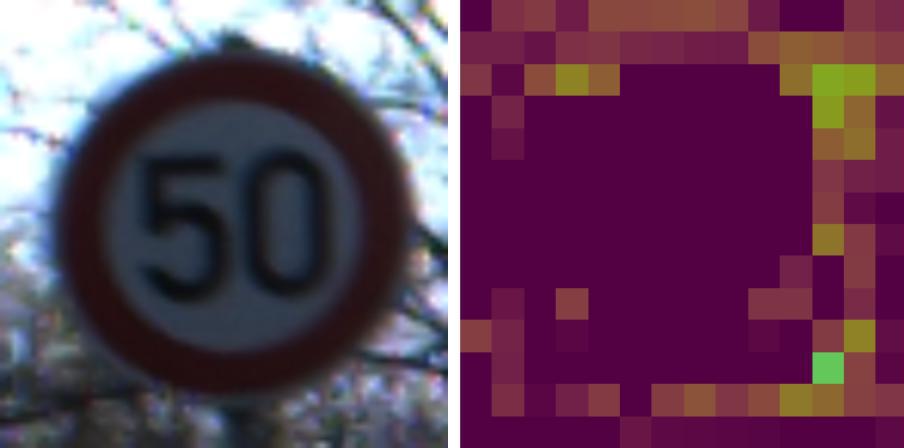}
        \end{subfigure}\\
        \begin{subfigure}[b]{\linewidth}
            \centering
            \includegraphics[width=\linewidth]{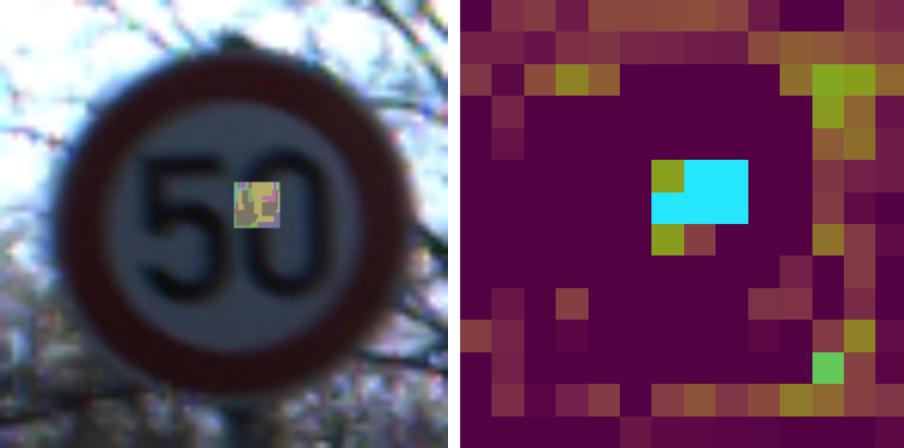}
        \end{subfigure}
        \caption{}
        \label{fig:defense_failure_5}
     \end{subfigure}

    \caption{Failed TPA defenses and occlusion maps. Figures~(a,b) are from PASCAL3D+ and~(c-e) are from GTSRB.}
    \label{fig:failed_defenses}
\end{figure*}

As shown in Figure~\ref{fig:roc}, these occlusion scores can be used to locate patches generated by either the TPA or Sparse-RS attacks. Note that this occlusion score does yield some false positives (if the model sees something in the image that does not conform to the ``object prototype"). False positives may arise because the model segments out the background, as in Figure~\ref{fig:stopsign}. Nevertheless, the occlusion scores achieve high recall at low thresholds, demonstrating that CompNets can successfully locate adversarial patches.

Occlusion maps also provide useful visualizations that explain how and why CompNets defend against patch attacks. In Figure~\ref{fig:successful_defenses}, we can see some cases where the CompNet detects the occluder and successfully defends against a patch attack for a variety of object classes. In Figure~\ref{fig:bus}, we see that model is able to detect and ignore patches even when they blend in. In Figure~\ref{fig:train}, the attack attempts to fool the model with semantically meaningful patches---wheels---but the CompNet recognizes them as occluders and ignores them. Moreover, the model can detect more than one patch well (Figures~\ref{fig:plane} and~\ref{fig:train}), backing up the quantitative results in Table~\ref{table:area_ablation}.

Even when the CompNet misclassifies the attacked image, the results are explainable. For instance, in Figure~\ref{fig:defense_failure_1}, we can see that the model has trouble determining the correct class of car because there is already significant occlusion present in the image (due to the signs on the car door and window). This is also seen in Figure~\ref{fig:defense_failure_2}, where it is hard to make out the main object, due to lack of contrast. Finally, the model may misclassify the image when salient and discriminative parts of the image are occluded: in Figure~\ref{fig:defense_failure_2}, the wheels are occluded, which may contribute to a misclassification. Moreover, in Figures~\ref{fig:defense_failure_3},~\ref{fig:defense_failure_4}, and~\ref{fig:defense_failure_5}, occluding a salient part of the traffic sign causes the model to confuse it for a similar class (even though it still detects the patches).

One practical advantage of this interpretability is that it allows us to better gauge our uncertainty about the model's predictions. In a real-world system, we could allow the model to predict an alternative ``uncertain" class whenever there is significant occlusion in the image. This could allow for graceful handling of adversarial patches and natural occluders, perhaps by deferring to other sensors or acting more cautiously. We hope that this advantage of compositional models is explored more thoroughly in future work.

    

\FloatBarrier
\section{Conclusion}
In this paper, we show that compositional representations are robust to patch attacks out of the box. Without expensive adversarial training, CompNets are able to detect, locate, and ignore adversarial patches. We confirm earlier findings that adversarial training of standard CNNs improves robustness to patch attacks, but show that adversarial training helps less than compositional representations. 
We also introduce part-based finetuning, a novel improvement to CompNets' training that boosts their fine-grained classification performance.
Finally, we show that CompNets' adversarial robustness is uniquely interpretable: their generative model can explain why a defense succeeds or fails. We are the first to demonstrate an architecture that can naturally resist patch-based attacks without adversarial training.

{\small
\bibliographystyle{ieee_fullname}
\bibliography{07_references}
}

\end{document}